\documentclass[runningheads]{llncs}
\usepackage[T1]{fontenc}
\usepackage{graphicx}
\usepackage{booktabs}
\usepackage[misc]{ifsym}
\newcommand{\corr}{(\Letter)}

\usepackage{researchpack}
\usepackage[dvipsnames]{xcolor}
\usepackage{hyperref}
\usepackage[capitalise,nameinlink]{cleveref}
\usepackage{booktabs}
\usepackage{enumitem}
\usepackage{xspace}
\usepackage{tikz}
\usetikzlibrary{positioning,shapes,arrows,calc}

\hypersetup{
    colorlinks=true,
    citecolor=blue,
    linkcolor=blue,
    filecolor=blue,
    urlcolor=blue,
}

\newcommand{\acronym}{Explanation-based Correction for \exstream\xspace}
\newcommand{\method}{{\tt ebc-exstream}\xspace}
\newcommand{\exstream}{{\tt exstream}\xspace}
\newcommand{\ddm}{{\tt ddm}\xspace}

\newcommand{\adwin}{{\tt adwin}\xspace}
\newcommand{\ph}{{\tt ph}\xspace}

\newcommand{\NB}{{\tt NB}\xspace}
\newcommand{\LR}{{\tt LR}\xspace}
\newcommand{\SVM}{{\tt SVM}\xspace}

\newcommand{\CSTAG}{{\tt c-stagger}\xspace}
\newcommand{\CELEC}{{\tt c-electricity}\xspace}
\begin{document}

\title{Spurious Correlations in Concept Drift: \\ Can Explanatory Interaction Help?}
\titlerunning{Spurious Correlations in Concept Drift}
\author{Cristiana Lalletti\inst{1} \and
Stefano Teso\inst{2,3}\corr}
\authorrunning{C. Lalletti and S. Teso}
\institute{Maths, University of Trento \and
CIMeC, University of Trento \and
DISI, University of Trento \\
\email{name.surname@unitn.it}}

\maketitle              % typeset the header of the contribution

\begin{abstract}
    Long-running machine learning models face the issue of \textit{concept drift} (CD), whereby the data distribution changes over time, compromising prediction performance.  Updating the model requires detecting drift by monitoring the data and/or the model for unexpected changes.
    We show that, however, \textit{spurious correlations} (SCs) can spoil the statistics tracked by detection algorithms.
    Motivated by this, we introduce \method, a novel detector that leverages model explanations to identify potential SCs and human feedback to correct for them.
    % It leverages an entropy-based heuristic to reduce the amount of necessary feedback, cutting annotation costs.
    %
    Our preliminary experiments on artificially confounded data highlight the promise of \method for reducing the impact of SCs on detection.
    
    \keywords{Concept Drift \and Spurious Correlations \and Confounding \and Interactive Machine Learning \and Human-in-the-loop \and Explainable AI.}
\end{abstract}

\section{Introduction}

In long-running machine learning applications, the data may undergo \textit{\textbf{concept drift}} (CD) \cite{gama2004learning}, whereby the distribution of observed inputs and associated outputs shifts, compromising the reliability of models learned on past data.
In order to update the (potentially obsolete) model after drift events, a variety of \textit{detection} strategies have been developed \cite{gama2004learning,gama2014survey,lu2018learning}.  These monitor the data distribution and/or the model's behavior and raise an alarm whenever these change more than would be expected by chance.

One issue that has so far been neglected is that of \textit{\textbf{spurious correlations}} (SCs) \cite{ye2024spurious,geirhos2020shortcut}:  these appear when unobserved confounders \cite{pearl2009causality} correlate the desired outcome and unrelated (non-causal) input variables, tricking learned models into wrongly relying on the latter.  Prototypical examples include background information \cite{beery2018recognition,schramowski2020making} and class-specific watermarks \cite{lapuschkin2019unmasking}.  Spurious correlations compromise generalization in even high-stakes applications like medical diagnosis \cite{geirhos2020shortcut} and scientific analysis \cite{schramowski2020making}.

Our key observation is that \textit{confounding skews the statistics that CD algorithms rely on}, hindering accurate and timely detection, yet current detectors are not designed to mitigate the impact of SCs.
To tackle this issue, we introduce \method (\acronym), a novel approach for drift detection that leverages \textit{machine explanations} and \textit{interaction} with a human expert.
\method belongs to the \exstream family of algorithms, which detect drift by monitoring the model's explanations, but it also includes an interactive step in which an expert user is asked to identify potential spurious correlations.  This feedback is then used by \method to \textit{deconfound} the model, facilitating timely identification of drift in its explanations and preventing relapse.
Moreover, \method minimizes cognitive load by implementing an entropy-based criterion to determine whether the model might affected by confounding and consult the user.

\textbf{Contributions}.  We:
1) Highlight a neglected interaction between SCs and drift detection, and show that it can affect CD detectors in practice;
2) Introduce \method, an explanation-based detector that obtains feedback from a domain expert to avoid this issue;
3) Report a preliminary evaluation of \method on an artificially confounded data, showcasing its promise.

\section{Preliminaries}
\label{sec:preliminaries}

\textbf{Setting}.  In online learning, a machine receives a sequence of examples $(\vx_t, y_t)$, for $t = 1, 2, \ldots$, where $\vx_t$ are observations and $y_t$ are labels.  The data are drawn independently from a time-varying distribution $p_t^*(\vX, Y)$.\footnote{We assume the labels to be available upfront, for simplicity.  Whenever they are not, they may be acquired using online active learning \cite{beygelzimer2009importance}.}
\textit{\textbf{Concept drift}} (CD) refers to situations in which this changes substantially, \eg $p_{t-1}^*(\vX, Y) \not\equiv p_t^*(\vX, Y)$, making learned knowledge obsolete and compromising prediction performance.
The problem is that of learning and maintaining a predictor $f: \vx \mapsto y$ that performs well on unseen instances \textit{even if the data distribution changes}.

\vspace{1em}

\noindent
\textbf{Dealing with drift}.  Strategies for handling CD can be broken down into two steps \cite{gama2014survey}:
\textit{i}) \textit{\textbf{detection}} involves monitoring the data distribution and/or the model for significant changes;
\textit{ii}) \textit{\textbf{adaptation}} involves updating the model appropriatey when drift is detected.
Naturally, it is essential that detection occurs as early and as accurately as possible.

A number of detection approaches have been proposed, see \cite{lu2018learning} for a recent survey.
Existing algorithms can be grouped into two main families.
Algorithms in the first one, like {\tt flora} \cite{widmer1996learning} and {\tt adwin} \cite{bifet2007learning} detect drift by monitoring the \textit{\textbf{data}}.
For instance, {\tt adwin} maintains a window of past examples and employs a statistical test to determine whether its distribution has diverged from that of current data, and triggers an alarm when this is the case.
Upon successful detection, the reference window is filled with post-drift data.

Approaches in the second family instead monitor for statistically unlikely changes in the \textit{\textbf{model}}.
For instance, \ddm \cite{gama2004learning} explicitly models the accuracy of the model as a Bernoulli random variable.  As long as the data are IID, the confidence interval of its parameter should improve \cite{valiant1984theory}, hence {\tt ddm} signals a drift event whenever this condition fails.
\exstream \cite{demvsar2018detecting} instead periodically computes (using {\tt IME} \cite{strumbelj2009explaining}) attribute-based explanations \cite{guidotti2018survey} of the model's decisions: these capture the relative importance of different input variables for a prediction, thus providing a succinct portrait of the model's decision process.  Then, it evaluates the dissimilarity of past and current explanations and pipes it into a baseline change detection algorithm, such as \ddm or \adwin.  Its main advantage is that differences between explanations are interpretable, and as such can be used to communicate to stakeholders the reasons \textit{why} an alarm was raised.

\section{The Impact of Spurious Correlations on Drift Detection}
\label{sec:spurious-correlations}

\textit{\textbf{Spurious correlations}} (SCs) occur whenever the training examples can be classified correctly using features that are not causally related to the label \cite{ye2024spurious}.
Formally, we assume that examples $(\vx_t, y_t) \sim p_t^*(\vX, Y \mid C)$ are influenced by a \textit{hidden} confounding variable $C$ that induces one or more SCs between $\vX_t$ and $Y_t$.
In image classification, for instance, the location in which a picture is taken can be a confound, as it influences both the context and labels of images (\eg wolves are more likely to appear on a snowy background than husky dogs \cite{ribeiro2016should}).
Due to factors like simplicity bias \cite{shah2020pitfalls}, predictors learned on confounded data favor learning such SCs over causally relevant features (\eg they tend to use the background to infer the label rather than the animal itself).
Since SCs affect the learned mapping between inputs and outputs, confounded predictors fail to generalize to unconfounded data.

\vspace{1em}

\noindent
\textbf{Impact on detection}.  These considerations beg the question: \textit{what is the effect of SCs on the behavior of drift detection algorithms?}
Under confounding, we are effectively dealing with \textit{two} potentially very different distributions:
the \textit{ground-truth distribution} $p_t^*(\vX, Y) \defeq \sum_c p_t^*(\vX, Y, C) $ that reflects how the world works and against which detection and model performance should be measured, and
the \textit{confounded distribution} $p_t^c(\vX, Y) \defeq p_t^*(\vX, Y \mid C = c)$ that examples are actually drawn from.\footnote{Throughout, we consider settings in which $C$ is independent of $t$.  We will briefly discuss other scenarios in \cref{sec:outro}.}
As a consequence, SCs can affect all detectors:
\begin{itemize}

    \item[(a)] Even if the ground-truth distribution changes under drift, by definition, yielding a detectable difference between $p_{t+1}^*$ and $p_t^*$, the difference between between $p_{t+1}^c$ and $p_t^c$ can be more modest.  This means that, depending on the choice of statistical test and divergence, \textit{data-based detectors} could fail to identify drift events on confounded data.

    \item[(b)] Models trained on confounded data tend to learn SCs rather than truly causal dependencies, meaning that the model's parameters, explanations, and performance are compromised as well, effectively masking drift events from model-based detectors.  For instance, the error rate of a model that strongly relies on an SC will remain roughly constant as long as this SC is in place, regardless of drift events affecting the other features.

\end{itemize}
In short, SCs negatively impact the ability of detecting drift events, regardless of the detector used, and existing detectors are not equipped for dealing with them.

SCs affect especially the model's explanations:  in our running example, an explanation would show that the classifier is in fact focusing on the background when predicting the label.  We will exploit this phenomenon to correct for confounding, as discussed next.

\section{Addressing Spurious Correlations in Drift Detection}
\label{sec:method}

We introduce \method, a detector that reduces the impact of SCs by eliciting corrective feedback for the machine's explanations.
\method works similarly to \exstream:  it monitors for changes to the model's explanations of predictions $\hat y_t$ for incoming inputs $\vx_t$ at regular intervals.  To this end, it invokes the {\tt SHAP} model-agnostic explainer \cite{lundberg2017unified}, and obtains explanations that describe the relative contribution of each input feature to the prediction.  Then it computes a dissimilarity between the current and reference explanations and pipes it through a regular drift detection algorithm like \ddm or \adwin.

In contrast to \exstream, however, it also: \textit{i)} applies a simple heuristic to check whether the model is affected by SCs, and \textit{ii}) if so, it asks a user to identify possible SCs in the model's explanations.
The second step consists of presenting the user with an explanation for the most recent input, and asking the user to indicate whether any of the features used by the model are in fact spurious.  For each of them, \exstream augments the training set by randomizing the spurious features, as done in \cite{teso2019explanatory}, thus reducing the label's correlation with them and forcing the updated model not to focus on them.

The first step leverages checks for confounding by computing the Shannon entropy of the (normalized) relevance weights of the most recent explanation and comparing it against a threshold:  low entropy indicates that the model is focusing on a small subset of features, which is a symptom of confounding.  This heuristic helps to minimizes cognitive load.

\section{Experiments}
\label{sec:experiments}

We present a preliminary empirical investigation of the following research questions:
\textbf{Q1}: Do SCs affect the operation of drift detectors?
\textbf{Q2}: Does \method help alleviate them?

\vspace{1em}

\noindent
\textbf{Algorithms}.  We compare \method with \exstream \cite{demvsar2018detecting}, using \ddm \cite{gama2004learning},\footnote{The \ph variant is indicated simply as \method or \exstream in the plots.}
\adwin \cite{bifet2007learning} or Page-Hinkley (\ph) \cite{page1954continuous,sebastiao2017supporting} as baseline detector, and experiment with three simple architectures trained in an online fashion:  a na\"ive Bayes classifier (\NB), logistic regression (\LR), and a linear support vector machine (\SVM).
Our code relies on the \texttt{river} library \cite{montiel2021river} and uses {\tt SHAP} \cite{lundberg2017unified} for extracting explanations.
All hyper-parameters are either fixed to their default values, or selected on a held-out \textit{confounded} validation set.

\vspace{1em}

\noindent
\textbf{Data sets}.  We manually inject SCs in two well-known data sets.
{\tt \underline{sta}gg\underline{er}} \cite{schlimmer1986incremental} requires classifying images of simple primitives with different {\tt shape} ($circle$, $square$, $triangle$), {\tt color} ($red$, $blue$, $green$) and {\tt size} ($small$, $medium$, $large$) based on whethery they satisfy one of three possible logical formulas:
$\varphi_1 = ({\tt size} = small) \land ({\tt color} = red)$,
$\varphi_2 = ({\tt color} = green) \lor ({\tt shape} = square)$, and
$\varphi_3 = ({\tt size} = medium) \lor ({\tt size} = large)$.
The training stream comprises $40,000$ examples and drift is induced every $10,000$ time steps by switching between formulas, yielding three drift events.
{\tt \underline{electricit}y} \cite{electricity} requires to infer whether electricity prices in the Australian electricity markets increase or decrease given $6$ features (time of the day, electricity price and demand in New South Wales and Victoria, and scheduled electricity transfer between these states).  It includes $45,312$ examples, collected from 7 May 1996 to 5 December 1998, with every instance covering a 30 minute window.  We estimated gold standard drift times using by running the {\tt HDDM} detector \cite{frias2014online}, obtaining three likely events at $t = 3,109; 18,712; 37,187$.
We construct confounded variants {\tt \underline{c-sta}gg\underline{er}} and {\tt \underline{c-electricit}y} by setting the label to $1$ whenever {\tt color} is \textit{green} or {\tt shape} is \textit{square}, encouraging the model to ignore {\tt size} for \CSTAG; or New South Wales electricity demand is larger than $0.45$, thus enforcing a correlation between this (already highly correlated) feature and the label for \CELEC.

\begin{figure}[!t]
    \centering
    \begin{tabular}{ccc}
        \includegraphics[height=9em]{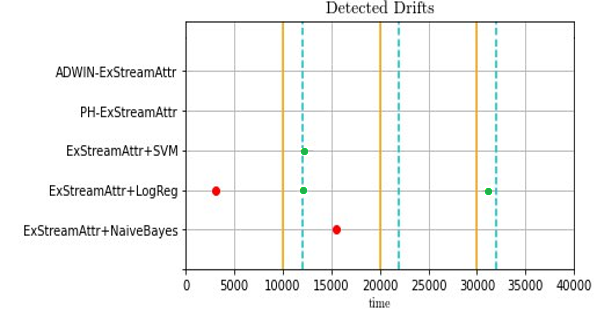}
        & & \includegraphics[height=9em]{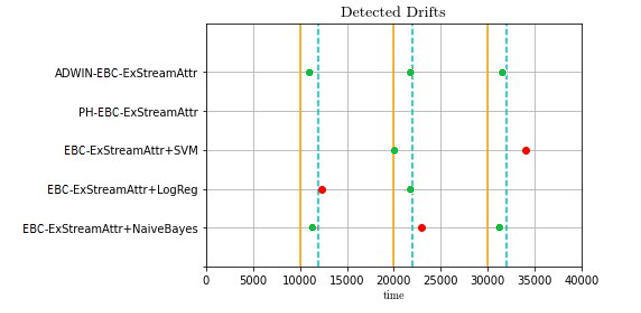}
        \\
        \includegraphics[height=9em]{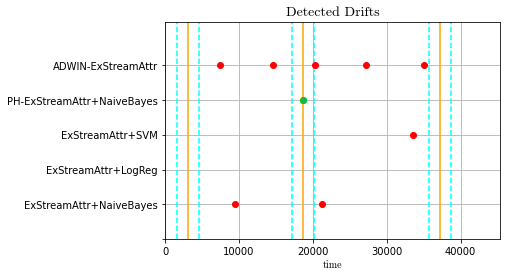}
        & & \includegraphics[height=9em]{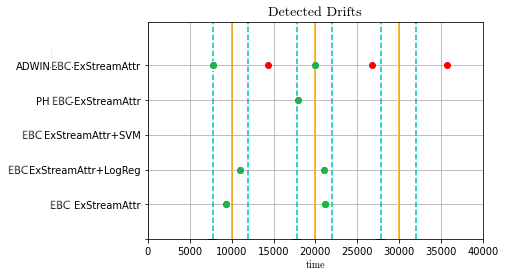}
    \end{tabular}
    \caption{Alarms raised by \method (right) and \exstream (left) on \CSTAG (top) and \CELEC (bottom).    Time is the $x$-axis, yellow bars denote ground-truth drift events, cyan dashed bars a heuristic acceptable delay.}
    \label{fig:results}
\end{figure}

\vspace{1em}

\noindent
\textbf{Results and discussion.}  The results, reported in \cref{fig:results}, show that \exstream either under-reports or over-reports drift events on both data sets.  For instance, \exstream detects almost no drift in \CSTAG (one \textit{late} drift for \NB when paired with \ddm, one drift for \SVM, and three drifts with one false positive for \LR).  In \CELEC the situation is similar, except that when used together with \adwin, \exstream tends to dramatically over-report, because in order to compensate for its lack of sensitivity -- due to SCs -- we had to lower the alarm threshold substantially, leading to false alarms.  \method, while still not perfect, fares much better, managing to uncover several more drift events in time (green dots).  Thanks to the entropy-based heuristic, the annotation cost is rather limited: for \CSTAG, \method elicits feedback at most $13-106$ times (over $40,000$ time steps) and for \CELEC at most $5$ times.

\section{Related Work}
\label{sec:related-work}

\noindent
\textbf{Spurious correlations in changing environments}.  To the best of our knowledge, the issue of SCs has been so far ignored in the CD literature, and existing detectors are not designed to be robust to SCs.
There exists work on other data quality issues in CD, such as contamination (\eg \cite{bhatt2022offline}) whereby examples are a \textit{mixture} of clean and corrupted observations.  SCs differ in that \textit{all} data is confounded.  Moreover, these approaches often assume an offline setup in which the data set can be segmented in a post-hoc fashion.  We make no such assumption.

Some works have studied how to build \textit{explainable} detectors capable of communicating \textit{why} they raised an alarm to users \cite{demvsar2018detecting,kulinski2020feature,budhathoki2021did,adams2021framework,hinder2023model}.  The \exstream algorithm, on which \method builds, is part of these efforts.  While these algorithms \textit{could} be used to interact with users for the purpose of deconfounding the detection process, as we do, in practice they neglect the issue of SCs and do not attempt to identify or fix it.
Recently, machine descriptions of detected drift have been used to interact with users \cite{bontempelli2022human} for correcting the machine's internal knowledge, but not for avoiding SCs.

\vspace{1em}

\noindent
\textbf{Handling spurious correlations}.  The issue of SCs is well known in machine learning and a variety of mitigation strategies have been proposed \cite{ye2024spurious}.
Our work is inspired by explanatory interactive learning techniques~\cite{kulesza2015principles,teso2019explanatory,schramowski2020making}, which acquire user corrections for machine explanations and encourage the model to ignore correlations annotated as spurious by the user.
These works, however, are unconcerned with online learning and drift detection \cite{teso2023leveraging}.

\section{Conclusion}
\label{sec:outro}

We have highlighted the issue of spurious correlations for drift detection, showing that SCs can hinder identification of drift events and, consequently, compromise future prediction performance.
Our preliminary experiments show that one can leverage machine explanations and human feedback to ameliorate the situation.

In future work, we plan to extend our analysis to more complex scenarios and machine learning models.
Specifically, while we consider only simple time-invariant confounds, in practice confounding can be more complex and change over time.  In this latter case, the user's feedback itself might become obsolete, requiring the machine to either acquire more or adapt the available feedback.  We leave an analysis of more challenging scenarios to future work.
Other possible research directions include extensions to self-explainable models capable of generating faithful explanations of their behavior \cite{teso2019toward,koh2020concept,marconato2022glancenets} and to add support for beneficial SCs \cite{eastwood2023spuriosity}.

\begin{credits}
\subsubsection{\ackname}
Funded by the European Union. The views and opinions expressed are however those of the author(s) only and do not necessarily reflect those of the European Union, the European Health and Digital Executive Agency (HaDEA) or the European Research Executive Agency. Neither the European Union nor the granting authority can be held responsible for them. Grant Agreement no. 101120763 - TANGO.

\subsubsection{\discintname}
The authors have no competing interests to declare that are relevant to the content of this article.
\end{credits}

\bibliographystyle{splncs04}
\bibliography{explanatory-supervision, references}
\end{document}